\documentclass[lettersize,journal]{IEEEtran}
\usepackage{amsmath,amsfonts}
\usepackage{algorithmic}
\usepackage{algorithm}
\usepackage{array}
\usepackage[caption=false,font=footnotesize,labelfont=rm,textfont=rm]{subfig}
\usepackage{textcomp}
\usepackage{stfloats}
\usepackage{url}
\usepackage{verbatim}
\usepackage{graphicx}
\usepackage{cite}
\usepackage{pifont}
\usepackage{booktabs}
\usepackage{multirow}
\usepackage{threeparttable}
\usepackage{rotating}
\usepackage{hyperref}
\usepackage{color}
\hypersetup{colorlinks = true,  linkcolor =blue,  anchorcolor = red,  citecolor = blue,  urlcolor = blue}

\begin{document}
	
\title{DVRP-MHSI: Dynamic Visualization Research Platform for Multimodal Human-Swarm Interaction}
%Shared Control based on Guiding Vector Fields \\ for Human-multirobot cooperation
\author{Pengming Zhu*, Zhiwen Zeng*, Weijia Yao, Wei Dai,  Huimin Lu, and Zongtan Zhou
% <-this % stops a space
%\thanks{* These two authors contributed equally to this work.}
\thanks{This work was supported by the National Science Foundation of China under Grant 62203460 and U22A2059; Major Project of the Natural Science Foundation of Hunan Province (No. 2021JC0004).}% <-this % stops a space
\thanks{* These authors contributed equally to this work.}
\thanks{Pengming Zhu, Zhiwen Zeng, Wei Dai, Huimin Lu, and Zongtan Zhou are with the College of Intelligence Science and Technology, National University of Defense Technology, Changsha 410073, China (e-mail: zhupengming@nudt.edu.cn.com; zengzhiwen@nudt.edu.cn; weidai\_nudt@foxmail.com; lhmnew@nudt.edu.cn; narcz@163.com).}
\thanks{Weijia Yao is with the School of Robotics, Hunan University, Changsha 410012, China (e-mail: wjyao@hnu.edu.cn).}}
		
%\thanks{Manuscript received April 19, 2021; revised August 16, 2021.}}
	
	
	
\maketitle
\pagestyle{empty}  % no page number for the second and the later pages
\thispagestyle{empty} % no page number for the first page
\begin{abstract}
In recent years, there has been a significant amount of research on algorithms and control methods for distributed collaborative robots. However, the emergence of collective behavior in a swarm is still difficult to predict and control.
 Nevertheless, human interaction with the swarm helps render the swarm more predictable and controllable, as human operators can utilize intuition or knowledge that is not always available to the swarm.
Therefore, this paper designs the Dynamic Visualization Research Platform for Multimodal Human-Swarm Interaction (DVRP-MHSI), which is an innovative open system that can perform real-time dynamic visualization and is specifically designed to accommodate a multitude of interaction modalities (such as brain-computer, eye-tracking, electromyographic, and touch-based interfaces), thereby expediting progress in human-swarm interaction research.
Specifically, the platform consists of custom-made low-cost omnidirectional wheeled mobile robots, multitouch screens and two workstations. In particular, the mutitouch screens can recognize human gestures and the shapes of objects placed on them, and they can also dynamically render diverse scenes. One of the workstations processes communication information within robots and the other one implements human-robot interaction methods.
The development of DVRP-MHSI frees researchers from hardware or software details and allows them to focus on versatile swarm algorithms and human-swarm interaction methods without being limited to fixed scenarios, tasks, and interfaces.
%This paper elaborates on the design of the platform as well as compares it with several existing swarm platforms, for which a brief survey has also been provided. 
The effectiveness and potential of the platform for human-swarm interaction studies are validated by several demonstrative experiments.
			
\end{abstract}
	
	
\section{Introduction}
Over the last few decades, swarm robotics has emerged as an attractive research field where multiple robots interact with the environment at a low level, resulting in the emergence of swarm intelligence.
This collective behavior can allow robot swarms to address problems with flexibility, robustness, and scalability beyond the capabilities of a single robot \cite{nedjah2019review}.
Therefore, robotic swarm systems have great potential for various applications, such as search and rescue \cite{niroui2019deep}, forest fire prevention and control \cite{ghamry2016cooperative}, area exploration \cite{roy2021exploration}, source seeking \cite{al2019multi}, and navigation \cite{mcguire2019minimal}.

In most cases, robotic swarms are expected to operate autonomously. However, predicting and managing their behavior remains a challenge.
The presence of a human operator would be beneficial, even necessary, since the operator could: (1) recognize and correct potentially erroneous decision-making of the robot swarms; (2) have intuition or additional information unavailable to the robot swarms that can be utilized to improve performance; (3) take responsibility for critical decisions when the robot swarms perform high-risk tasks; (4) adjust the intention appropriately as the mission goals change.

While many human-swarm interaction (HSI) methods have been developed \cite{kolling2015human}, their practical implementation is still difficult and calls for further improvement.
Therefore, it is necessary to design a platform for HSI research to investigate how to develop more intuitive methods to control swarms and reduce the cognitive burden of human operators, which is the motivation of this paper.

In our opinion, an ideal HSI platform should have the following features:
(1) The robots should be low-cost, easy to manufacture and operate, effortless for maintenance, and with flexible motion capabilities; (2) This platform facilitates the deployment of various user-defined scenarios for testing various algorithms; (3) The platform can dynamically display various scenarios and visualize the robot swarm statuses to the user, thus facilitating participation in real-time interventions; (4) The platform is able to accept multiple types of input devices to explore more HSI modalities such as electroencephalogram (EEG), electromyographic (EMG) signals, eye tracking, gesture recognition, touch control, and voice recognition.
The development of such a platform would reduce the burden on researchers, allowing them to focus more on swarm algorithms and human-robot interaction study, free from the constraints of fixed scenarios, tasks, and interfaces.

\begin{figure}[t!]
	\centering
	\includegraphics[width=0.92\linewidth]{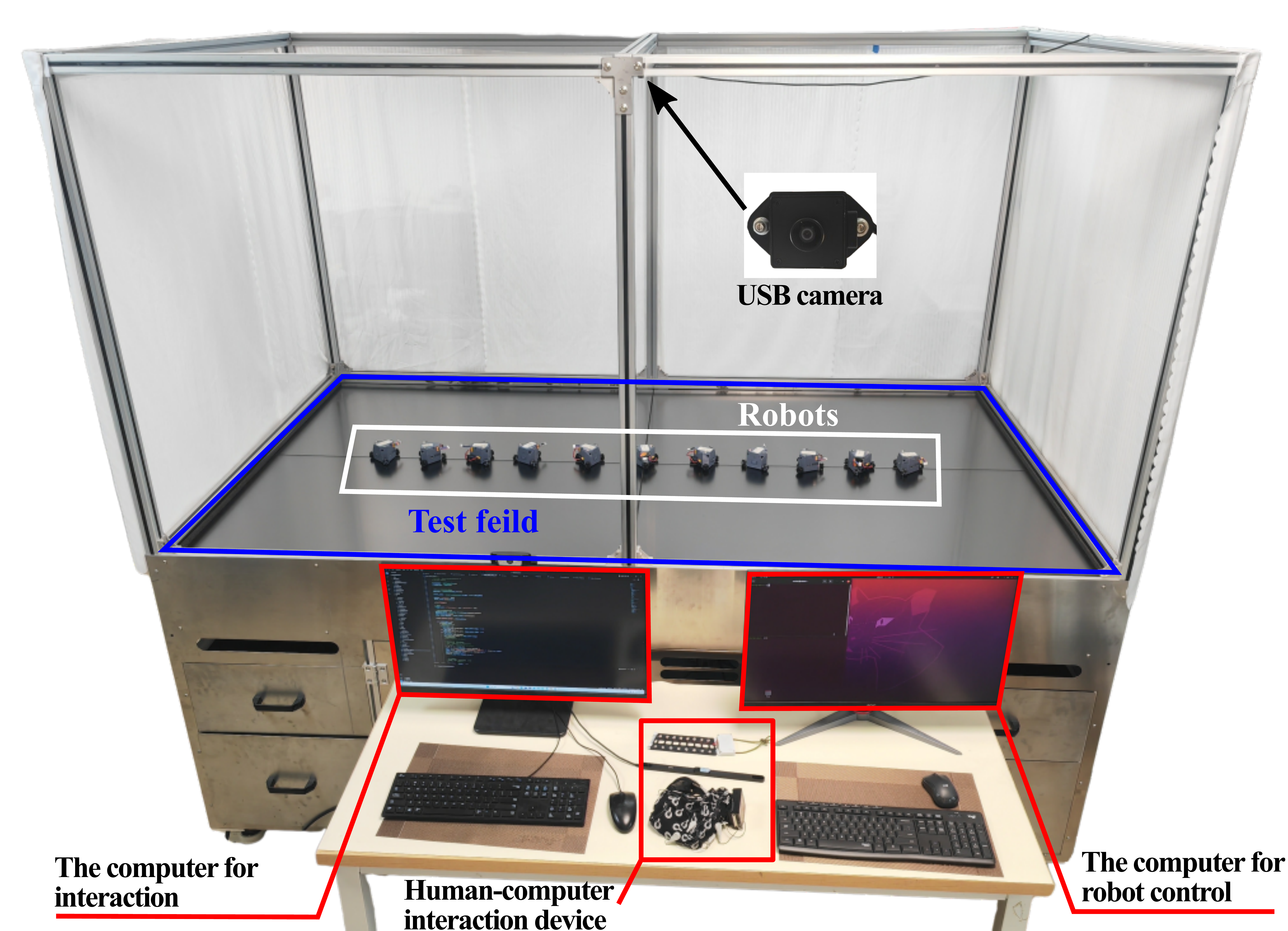}
	\caption{Full system testing field of DVRP-MHSI.} 
	\label{system}
\end{figure}

In this paper, we develop a dynamic visualization research platform for multimodal human-swarm interaction (DVRP-MHSI), as shown in Fig. \ref{system}. This platform allows future HSI research on arbitrary swarm algorithms, in various scenarios, and with multiple modal inputs.
The main contributions of this paper are as follows:
\begin{itemize}
	\item [1)]
	We develop a dynamic display platform based on multitouch screens, which can directly acquire human touch points and recognize robots and objects placed on them. It allows users to set up arbitrary task scenarios and display various robot states.
	\item [2)]
	We devise a small three-wheeled omnidirectional wheeled mobile robot that is low-cost and easy to maintain, and its chassis is entirely 3D printed. Benefiting from its special structure, the self-designed robot can move flexibly and omnidirectionally.
	\item [3)]
	We design a flexible software architecture that allows users to easily modify communication connections, inputs, outputs, and custom parameters. Following our protocols, various human-machine interfaces can be conveniently integrated into our system.
\end{itemize}

The rest of the paper is organized as follows.
Related work on human-swarm systems is introduced in Section \uppercase\expandafter{\romannumeral2}. 
Section \uppercase\expandafter{\romannumeral3} explains in detail the design of the components of DVRP-MHSI.
The results of several demonstration experiments are presented and discussed in Section \uppercase\expandafter{\romannumeral4}.
%Finally, Section \uppercase\expandafter{\romannumeral5} presents the conclusion and the future work for the DVRP-MHSI system development.
Finally, Section \uppercase\expandafter{\romannumeral5} presents the conclusion for the DVRP-MHSI system development.

\section{Related work on swarm systems}
In the past decades, a considerable number of swarm robotic platforms have been developed to meet different research needs\footnote{A brief survey has been provided in the supplement to this paper at \href{https://github.com/pengming-nubot/Supplement-to-DVRP-MHSI}{https://github.com/pengming-nubot/Supplement-to-DVRP-MHSI} }.
%This section introduces existing simulators, hardware platforms, and human-swarm platforms for swarm research.
This section introduces existing human-swarm platforms for swarm research.
Theoretically, a human-swarm system can be constituted by directly introducing human intervention into the robotic swarm system.
However, some platforms, such as Kilobot \cite{rubenstein2014kilobot}, Jasmine \cite{kernbach2009re-embodiment}, and Alice \cite{caprari2003design}, are decentralized with no central computer to facilitate communication. This implies that each robot is limited by its communication and sensing capabilities, which is not conducive to human-robot interface access.
Therefore, robots should have some perception, communication, and decision-making capabilities, and thus a simulated distributed control approach can be used.
A simulated decentralized hardware swarm is defined as a platform with a central computer containing specific information and action rules for each robot, allowing the researcher to selectively distribute information about the environment to individual robots \cite{dhanaraj2019adaptable}.
While such a system is centralized, researchers can implement decentralized algorithms by simulating the robot's behavioral rules and controlling the data it receives.

Using GRITSBot \cite{pickem2015gritsbot}, the swarm platform Robotarium \cite{pickem2017robotarium,wilson2020robotarium} is built according to the simulated distributed approach.
Robotarium is an inexpensive, open-source platform that lowers the barrier to entry for swarm research, allowing researchers to conduct experiments remotely. 
The system allows easy transfer from simulation to hardware and includes an interface for interaction and data collection, providing the foundation for human-swarm interaction research.
Similarly, the APIS \cite{dhanaraj2019adaptable} platform also employs a simulated decentralized design that enables simple control of small autonomous robots that can easily interact with humans as a swarm desktop interface, and a preliminary case of human-swarm interaction was tested using peripheral inputs.
An augmented reality interface is designed as an application for a handheld device by Patel et al \cite{patel2019mixed-granularity}. Algorithm validation based on the simulator ARGoS \cite{pinciroli2012ARGoS} is performed before sending it to the robot for execution, enabling collaborative object transportation.
In addition, using E-puck \cite{mondada2009puck}, Podevijn et al. \cite{podevijn2016Investigating} organize experiments to study the effects of growing robot numbers on the human physiological state in human-machine swarm interactions.
Kim et al. \cite{kim2019user-defined} use several Zooids \cite{le2016zooids} to build a platform for human-swarm robot interaction and propose a user-defined interaction method for the swarm control of desktop robots.
The platform enables the control of the swarm robot through camera-recognized gestures, such as 1-2 fingers, one hand or both hands, etc.

Notably, almost all of the above platforms invariably use VICONs or cameras to acquire the robot's locations. This allows researchers to focus on algorithms and human-robot interaction studies without being stuck with robot localization.
Besides, simulating a decentralized robotic swarm allows users to access a variety of interfaces and obtain information about the entire swarm, thus providing the necessary flexibility for HSI and human-computer interface research.
Therefore, research can be conducted on the ability of different interaction modalities to ultimately promote an open dialogue between the user and the swarm.

%In contrast, working with on-table wheeled robots lets us deploy, without computer-rendered or VR simulation, relatively numerous groups of small robots (i.e., closer to future envisioned swarm systems) operating on a 2-dimensional workspace. This capability provides us with the unique opportunity to investigate the effects of group size and robot proximity on user input preference across a wide swath of example tasks, without worry that our results will suffer from the documented “reality gap” that exists in simulated (as opposed to implemented in hardware) human-robot interaction studies.

\begin{figure*}[h!]
	\centering
	\includegraphics[width=0.92\linewidth]{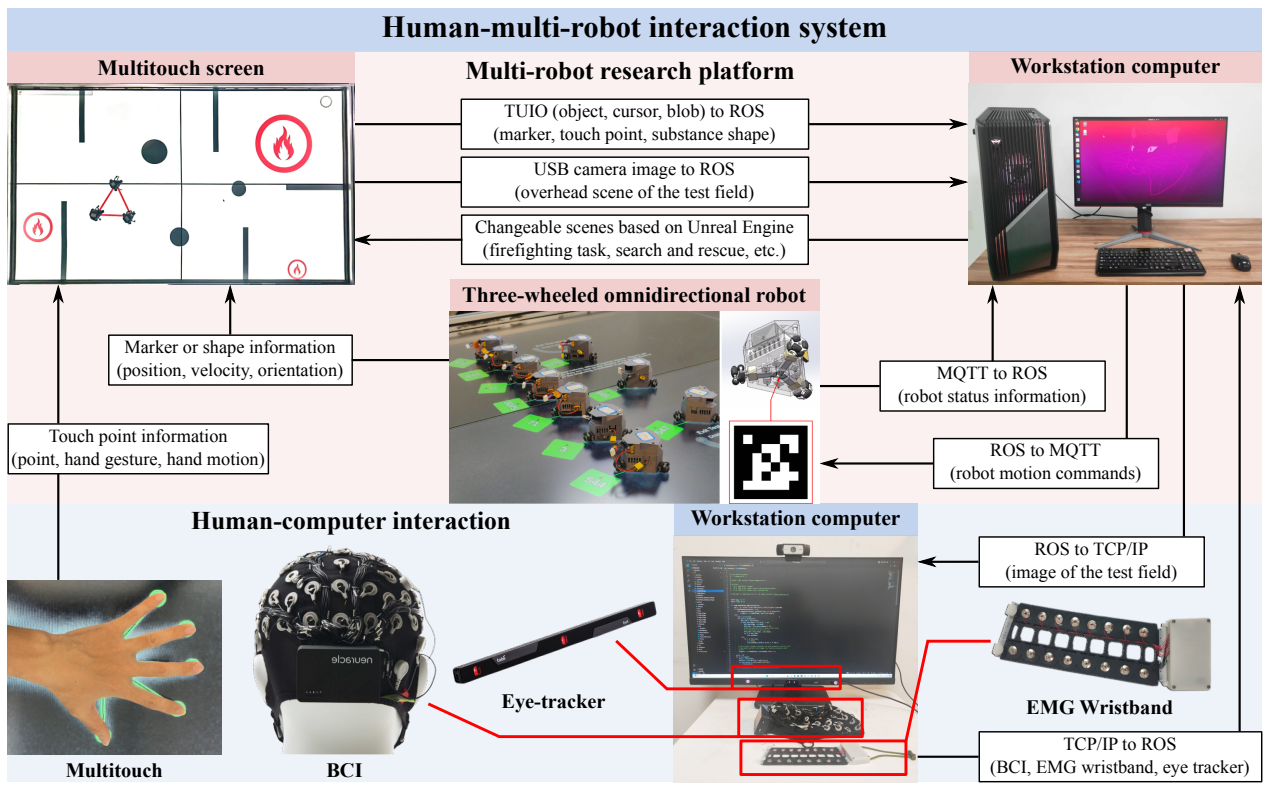}
	\caption{The designed human-multi-robot interaction system.} 
	\label{system framework}
\end{figure*}
\section{system design of DVRP-MHSI}
DVRP-MHSI is a testbed designed to accelerate the development of human-swarm interactions.
It allows access to multiple human-machine interfaces and versatile swarm algorithm development.
The design of this system meets the requirements of the envisaged ideal HSI platform as discussed in Section I.
The human-multi-robot interaction system consists of three main parts: the test field based on multitouch screens (named MultiTaction cell) and infrastructure, the three-wheeled omnidirectional robot\footnote{The details about the design of the robot (including structural design, electrical design, and motion control) and the results of the robot locomotion testing can be found in the supplement to this paper at \href{https://github.com/pengming-nubot/Supplement-to-DVRP-MHSI}{https://github.com/pengming-nubot/Supplement-to-DVRP-MHSI} } that comprises the swarm, and the software framework designed for interaction interfaces and robot control, as shown in Fig. \ref{system framework}.
In this section, we introduce the design of DVRP-MHSI's hardware and software architecture and the motivation behind it.
	
\subsection{Test field based on multitouch screens}
The test field of DVRP-MHSI, shown in Fig. \ref{system}, consists of four multitouch screens, a USB camera, and two system computers. 

The multitouch screens are the main testing ground for the robots to place, move, and perform various tests.
Human operators can directly interact with four multitouch screens, with the multitouch screen capturing human touch point information or gestures (position, velocity).
Moreover, the screens demonstrate their advanced capabilities when robots or other objects are placed on them. They can accurately recognize the shape of the robot (object) or the QR code affixed to the bottom, providing detailed information about the position, velocity, orientation, and acceleration of the robot (object), as shown in the robot part of Fig. \ref{system framework}.
Therefore, each robot does not need to be equipped with localization sensors.

%Then, the screen transfers all information to the computer in TUIO format, which includes cursor, object, and blob.
%On the workstation side, we can design arbitrary virtual scenarios based on Unreal Engine for certain tasks, for instance, typical search-and-rescue, fire-fighting scenarios, etc.
%Received data from the screens and robot status information in ROS message format, the computer runs the appropriate program.
%Then, it sends the correct commands to the robot in MQTT format, with changing elements in the scene.
%In addition, we can utilize more human-computer interactions such as brain-computer interface and eye-tracking device to express human intention, promoting more exploration of human-multi-robot collaboration methods.
%The three-wheeled omnidirectional robot with sensors, as the actuating part of the overall system, has excellent motion performance. 
%The robots communicate with the computer and other robots via a wireless network.
%, thus providing the basis for human-robot interaction and multi-robot collaboration.
%When the robot receives a command, it executes the corresponding action and provides real-time feedback to the computer about its status.
%Besides, an operator can place or remove any robot from the screen at anytime, meaning the robot joins in or withdraws from a group.
%
%The touchscreens provide information about the robots' positions and are simulated decentralized so that no localization sensors are mounted on each robot. 
The main motivations for using multitouch screens in the DVRP-MHSI are their ability to display the robot states dynamically in real time and to set up various experimental scenarios. They also have high accuracy in robot localization, with an average error of millimeters in the laboratory environment.

MultiTaction Cell\footnote{Some important parameters can be found in TABLE  \uppercase\expandafter{\romannumeral3} in the supplement to this paper at \href{https://github.com/pengming-nubot/Supplement-to-DVRP-MHSI}{https://github.com/pengming-nubot/Supplement-to-DVRP-MHSI}} is used as our test screen, and it is a multi-touch modular LCD screen commonly used for, but not limited to, advertising, exhibition, and education.
%\ref{multitaction}.
%Multitouch means that Cells can track and react to several people interacting with them simultaneously.
%Moreover, the system tracks a person's hands instead of points of contact only, enhancing interactive possibilities.
%Modular means that Cells can be easily stacked and combined to form a video wall, showing interactive content and reacting to multiple users' touch.
In our system, four Cells are stacked to form a single large display array.
The tracking engine runs automatically on the Cell's internal computer. 
The Cell processes the image data captured by its camera matrix when the user touches or places an object on it.
Then, it outputs this information as tracking data, including fingertip locations, infrared pen locations, marker locations, etc.
In particular, the number of simultaneous touch inputs is unlimited, so the screen can theoretically recognize all objects placed above it if there are no space constraints.

The USB camera, a part of the DVRP-MHSI, is fixed directly above the screen and provides high-definition video to record the robot's execution during the experiment.
Using the OpenCV library, the camera corrects distortion and applies projection transform to ensure the captured footage aligns with the test field area.

In addition, the two system computers are part of the infrastructure of the DVRP-MHSI.
One of the computers is responsible for robot/subsystem communication, decision-making, control, and test environment simulation.
The other is responsible for accessing various human-computer interfaces, including the brain-computer interface, the electromyographic wristband, and the eye-tracking device.

\begin{figure*}[t!]
	\centering
	\includegraphics[width=\linewidth]{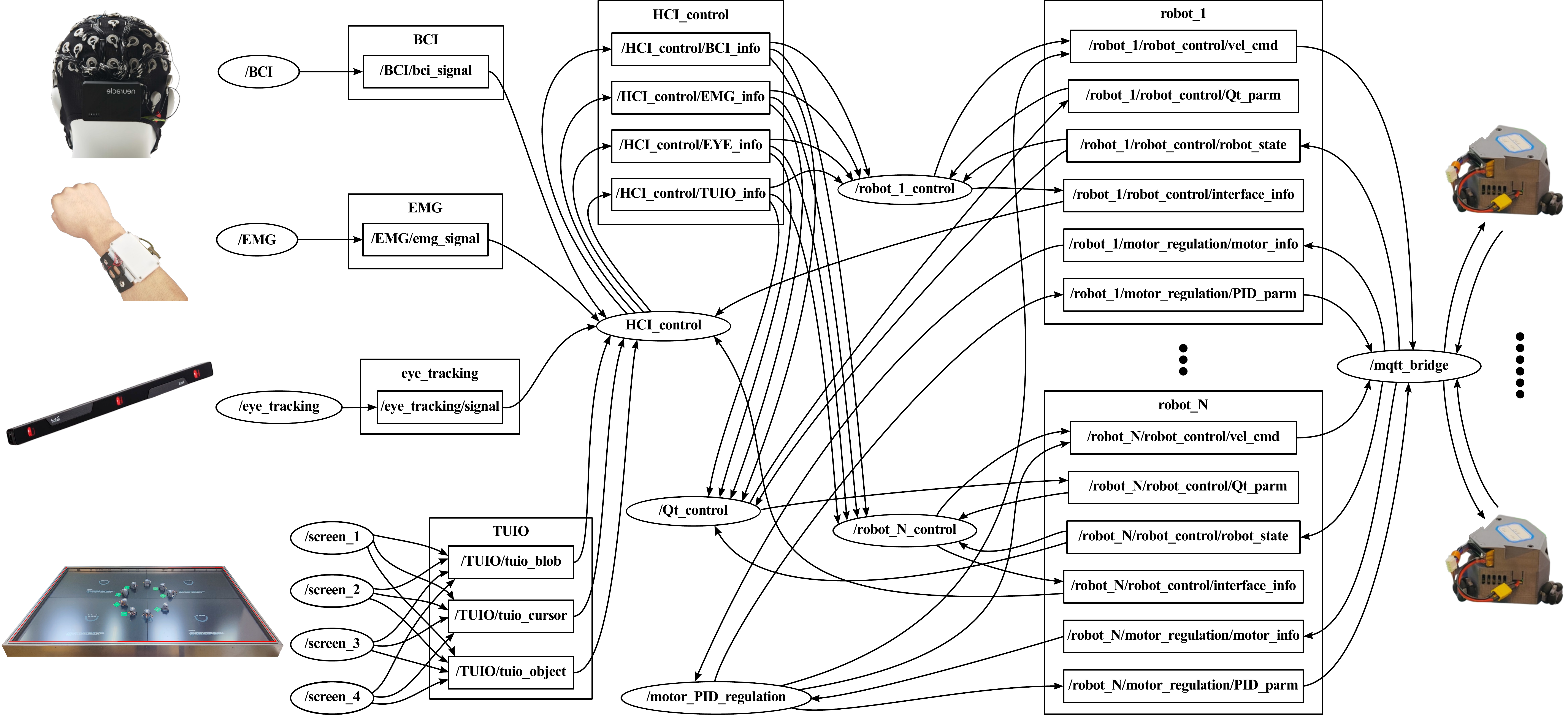}
	\caption{The software design of DVRP-MHSI.} 
	\label{software_design}
\end{figure*}

\subsection{Software design}

Alongside the design of the platform hardware, a software architecture has been developed to support HSI research. 
The objective is to accommodate a diverse range of robots, input devices, and task sets, emphasizing flexibility for future developments to focus on various HSI research.

The robot communicates with the workstation over a wireless network, using Robot Operating System (ROS) and Message Queuing Telemetry Transport (MQTT) to connect them, thus making it ROS-compatible.
ROS is an open-source meta-operating system for robotics applications. 
It provides services similar to those of a typical operating system, including hardware abstraction, underlying device control, implementation of common functions, inter-process messaging, and package management.
Its communication is carried out through a publish/subscribe model, in which topics consisting of predefined information structures can be communicated between multiple nodes (processes) in a network. These topics can be accessed by any node in the network, allowing for easy scalability of publishers and subscribers.

However, due to limited CPU resources, most swarm robots, including ours, cannot handle a full local ROS instance. 
To integrate these features on less powerful microcontrollers without a full ROS instance, we implement a communication module using the MQTT protocol, which has been shown to be a reliable and scalable communication method for swarm robotics systems \cite{efrain2018implementation}. 
MQTT is a communication protocol with a lightweight design and a small header that significantly reduces network traffic, which makes it particularly suitable for resource-limited IoT.
MQTT supports publish/subscribe messaging mode, allowing one-to-many message publishing, which can effectively decouple applications. 
It is also based on a TCP/IP network connection, which provides an orderly, lossless, bi-directional connection and ensures the reliable transmission of messages.
Through this communication method, the workstation can send/receive customized messages via ROS and connect to a wide range of ROS applications to allow message playback for debugging, thus facilitating algorithm development.

In addition, the user only needs to define the ROS message type and format of the interaction interface and communicate with the workstation via TCP/IP.
Therefore, various interaction methods can smoothly access the entire communication network.
It is worth noting that this approach allows multiple users/interactions to access simultaneously, with each user/interaction treated as an independent node.
On the robot side, MQTT enables lightweight communication, reducing the load on the central processor and enabling direct communication between robots within the network in a well-defined manner. 
Upon receiving a command, the robot executes the corresponding action and provides real-time feedback to the computer regarding its status.
%Besides, an operator can place or remove any robot from the screen at anytime, meaning the robot joins in or withdraws from a group.

\begin{figure}[h!]
	\centering
	\includegraphics[width=0.8\linewidth]{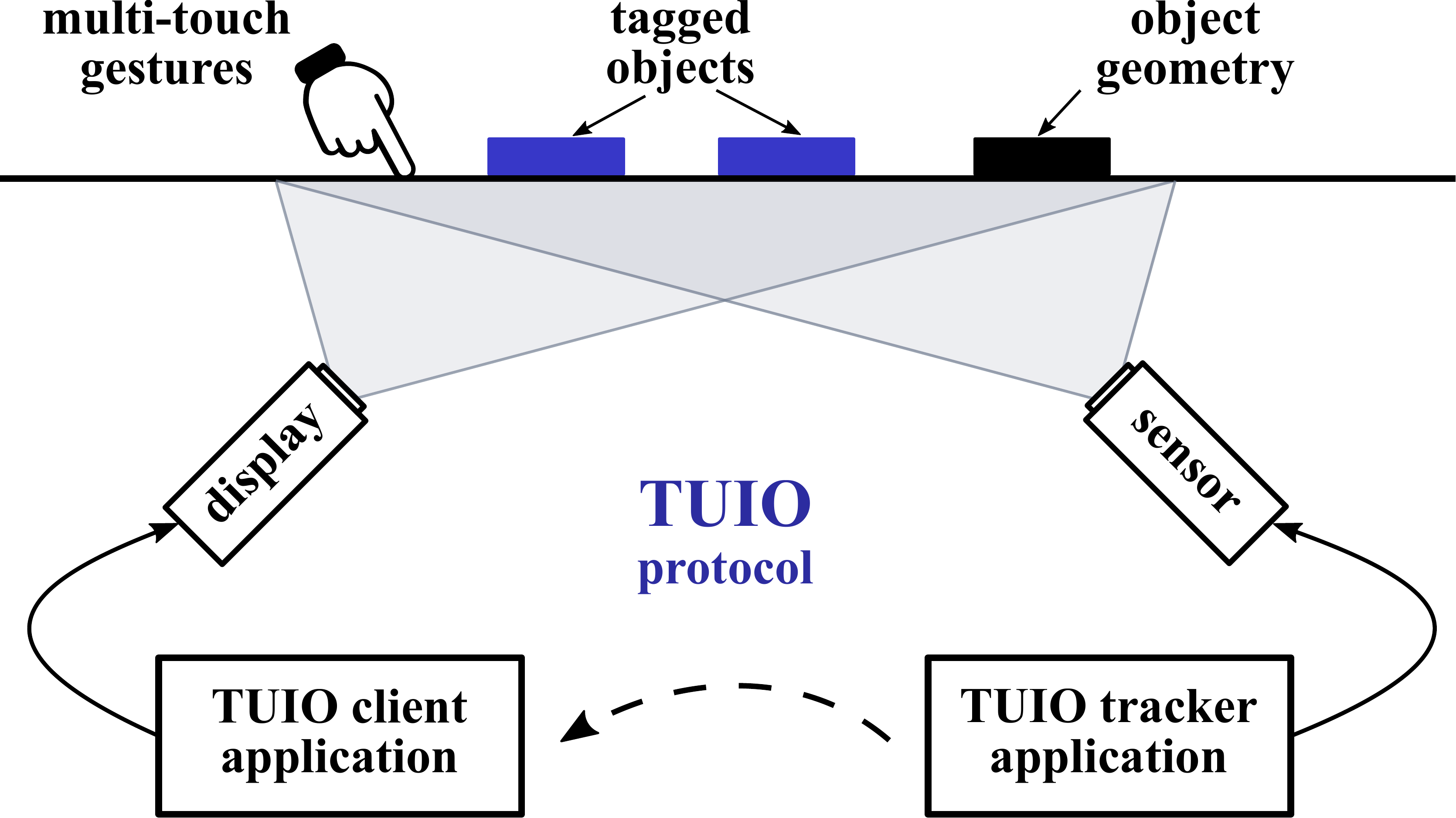}
	\caption{The illustration of TUIO protocol.} 
	\label{TUIO}
\end{figure}

The multitouch screen is connected to the workstation via Ethernet and transmits all acquired information to the workstation using the TUIO protocol. The principle of operation is illustrated in Fig. \ref{TUIO}. TUIO, a protocol based on Open Sound Control (OSC) for table-top tangible user interfaces, is widely used in touchscreen messaging and includes three main communication formats: cursors, objects, and blobs. The object types are mainly for objects with QR codes, including their position, velocity, angular velocity, and angular acceleration. The blob types are mainly for regular objects placed above, including their position, length, width, area, velocity, angular velocity, and angular acceleration. Similarly, this information is encapsulated in ROS message format for distribution to other nodes.

Additionally, virtual scenarios can be created using Unity to simulate various tasks, such as search-and-rescue operations or fire-fighting scenarios. The corresponding scenes are displayed on the screen, providing an interactive and immersive environment for these activities.

\subsection{Interaction method}
\begin{figure}[t!]
	\centering
	\includegraphics[width=0.9\linewidth]{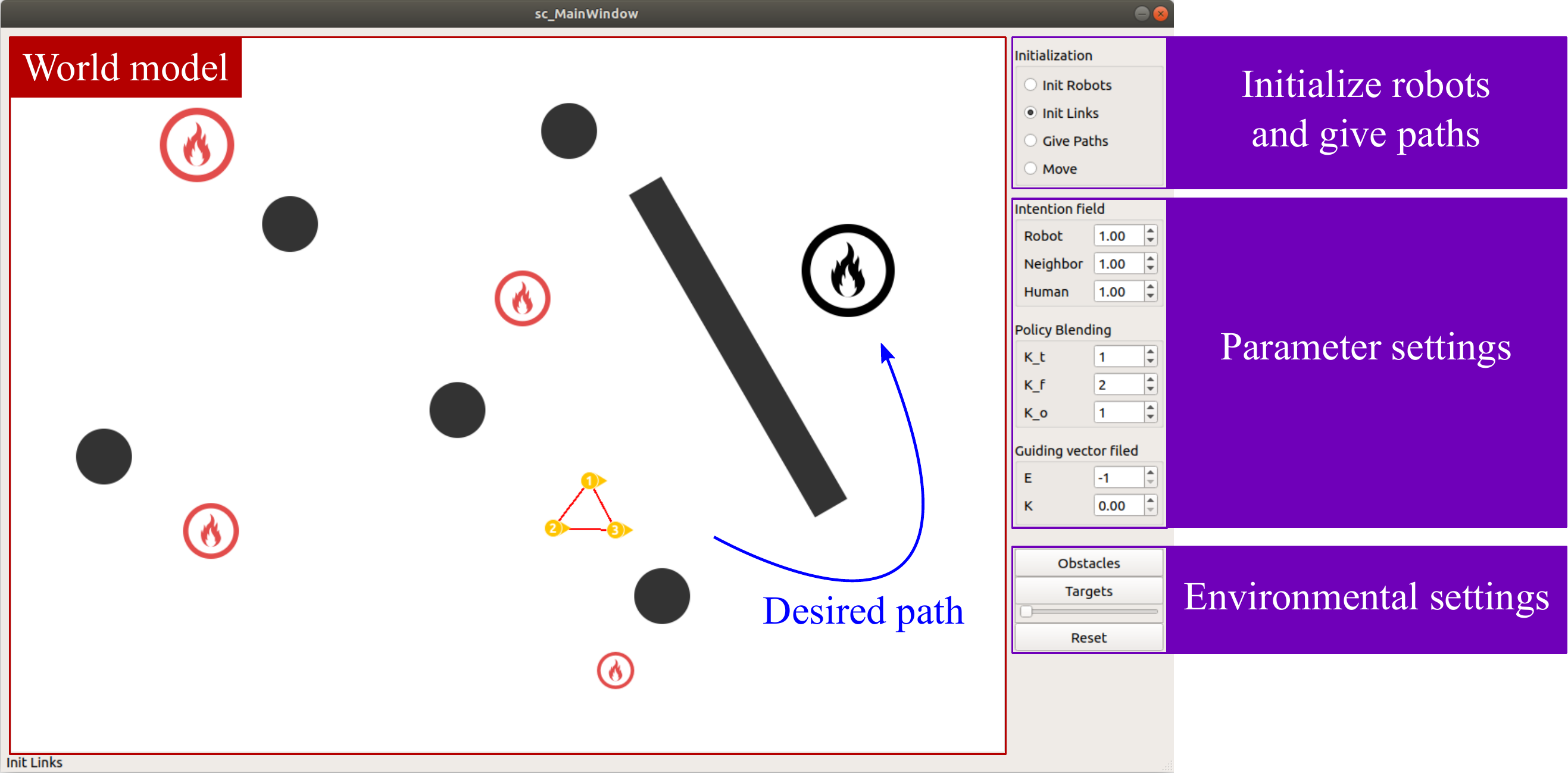}
	\caption{Designed QT interface.} 
	\label{QT}
\end{figure}

Thanks to the meticulous design of our software architecture, users can freely and flexibly employ various interaction methods by customizing individualized information, significantly enhancing the system's adaptability and user-friendliness.
To initially validate the effectiveness of the human-computer interaction approach, a human-computer interface based on QT is designed, as shown in Fig. \ref{QT}.

% The interface has a variety of application, as follows:
% \begin{itemize}
% 	\item [$\bullet$]
% \textbf{Monitoring Robot Status}: It functions as a console to monitor real-time status updates from the robot, including vital information such as location, velocity, and operational mode.
% 	\item [$\bullet$]
% \textbf{Scenario Design}: The interface allows users to design and configure multiple scenarios. These scenarios can be tailored for various tasks and simulations, such as testing different robot behaviors, evaluating responses to specific environmental conditions, or planning complex missions.
% 	\item [$\bullet$]
% \textbf{Human-Robot Interaction}:It facilitates interaction between humans and robots by displaying relevant information. Moreover, it can be integrated with various human-computer interaction methods to control and display corresponding human intentions.
% \item [$\bullet$]
% \textbf{Control and Command}: Depending on its capabilities, the interface may also provide controls for commanding the robot, setting waypoints, adjusting parameters, and initiating actions remotely.
% \end{itemize}
The interface is a comprehensive tool that provides real-time status monitoring, scenario design, human-robot interaction, and remote control capabilities. It enhances efficiency and user experience in various applications, from research and development to practical deployment scenarios.
Besides, information from other human-computer interfaces can also be transferred to this QT interface.
This section will introduce several typical human-computer interactions, such as the brain-computer interface, the electromyographic wristband, and the eye-tracker.

\subsubsection{Brain-computer interface}
The brain-computer interface (BCI) is a special human-computer interaction scheme that directly uses the brain's consciousness activity to exchange information with the external world \cite{wolpaw2002brain}.
Researchers typically use noninvasive electrode arrays placed along the scalp to capture varying degrees of electrical activity in the cerebral cortex.
In contrast to traditional interactions, the most important feature of BCI is the generation of intentions directly based on EEG signals, which reflects the inclination and uncertainty of human choice \cite{lotte2018areview}.
Therefore, BCI is an appropriate human-computer interface for controlling intelligent robots.
In our previous work, we investigated the classification processing of EEG \cite{li2024multi}, applying BCI techniques to the multi-robot field \cite{liu2021brain,dai2021shared}.
Our platform uses the NeuSen W wireless EEG acquisition system as BCI experimental equipment, as shown in Fig. \ref{system framework}.
This system is portable, signal-stable, and well-shielded. It is also equipped with a nine-axis motion sensor, allowing subjects to conduct traditional psychology/BCI research in the shielded room or more natural and flexible experiments outside the shielded room.
The hardware part of the acquisition system consists of a 64-lead EEG cap and an amplifier. The electrode positions are arranged according to the international 10-20 system. 
The EEG caps are available in three sizes to accommodate various head shapes, ensuring signal quality and comfort during acquisition.

% \begin{figure}[h!]
% 	\centering
% 	\subfloat[BCI]{\includegraphics[width=0.3\linewidth]{figure/BCI}}
% 	\label{BCI}
% 	\subfloat[EMG]{\includegraphics[width=0.3\linewidth]{figure/EMG}}
% 	\label{EMG}
% 	\subfloat[eye-tracker]{\includegraphics[width=0.3\linewidth]{figure/EYE}}
% 	\label{EYE}
% 	\caption{Typical human-machine interfaces to DVRP-MHSI.} 
% 	\label{interfaces}
% \end{figure}

\subsubsection{EMG interface}
The EMG interface captures the electromyography signals from the human epidermis through sensors to discern the human intent.
EMG signals typically have a bandwidth of around 1 kHz, with significant signal components concentrated between 0 and 500 Hz.
Muscle contraction potential differences measured at the body surface range from tens to hundreds of microvolts. 
Despite its weak signal strength and susceptibility to noise, EMG signals directly reflect body activity, making it an advanced human-computer interface for predicting user intention \cite{atzori2014electromyography}.
Moreover, there has been much work on controlling robots/swarms based on EMG.
%\cite{wolf2013gesture,kundu2018hand,luo2020a,suresh2020human}.
We have designed electromyography acquisition devices, as shown in Fig. \ref{system framework}.
An array of snap buttons is located on the inside of the wristband. The snap buttons are 3.9mm standard medical snap buttons, which can be matched with standard dry Ag/AgCl electrodes and gel electrodes.
The button array is arranged in pairs longitudinally along the arm, capturing distinct muscle activities.
Typically, a configuration involving two electrodes is aligned with the muscle fibers, forming a differential pair. This setup enables precise differential amplification of the surface EMG signals. Furthermore, these electrode pairs are evenly dispersed around the circumference of the arm, enhancing spatial resolution and signal diversity.
Our wristband can deliver eight distinct channels for surface EMG signal acquisition. This multi-channel design significantly augments the system's sensitivity and fidelity, empowering it to process even diminutive EMG signals effectively. Consequently, this wristband excels in discerning minute hand gestures and subtle motion nuances.

\subsubsection{Eye tracking}
The eye-tracker monitors and documents human ocular movements using infrared light and camera technology for data acquisition. This rich dataset can be analyzed to infer users' objectives, forecast future behaviors, and insights into their cognitive states \cite{koochaki2018predicting}.
Many studies have used eye-tracking techniques to recognize human intentions for indirect input or task assistance.
%\cite{shafiei2024development,aronson2018eye,ming2016anticipatory}. 
Furthermore, intent prediction through eye movement is natural and seamless and can be used as an efficient interface.
Our system uses a simple Tobii eye-tracker to capture human eye movement, as shown in Fig. \ref{system framework}. 
The eye-tracker is equipped with three infrared light sources and sends them outward at 33 HZ.
And there is a custom infrared optical sensor under the device's housing to pick up the infrared light reflected from the eyes to capture eye movements, with a sampling rate of up to 120 HZ.
In addition, the eye-tracker uses head-tracking algorithms to capture head movements more accurately.
Despite the non-commercial edition's current limitation to gaze positions, we can further analyze changes in gaze points over time. This analysis enables us to derive gaze trajectories and construct heat maps that illustrate attention distribution and estimate cognitive workload.

%\subsubsection{Touch control}

%\subsubsection{Voice recognition}
\section{Validation}
In this section, to demonstrate the efficacy of DVRP-MHSI, we conduct a series of experiments.
%including robot locomotion testing and human-multi-robot interaction.
Rather than focusing on the algorithms used, we demonstrate a preliminary implementation of human-robot interaction.
All participants were provided with written informed consent to participate in the experiment in accordance with the Declaration of Helsinki.

We test a variety of human-computer interaction methods for three different case studies: (1) BCI for multi-target selection, (2) EMG for generating robot action commands, and (3) eye tracking for desired trajectory generation.
%; (4) multitouch for target area and trajectory generation.
\subsection{BCI for multi-target selection}
\begin{figure}[t!]
	\centering
	\subfloat[The process of selecting targets with BCI]{\includegraphics[width=0.49\linewidth]{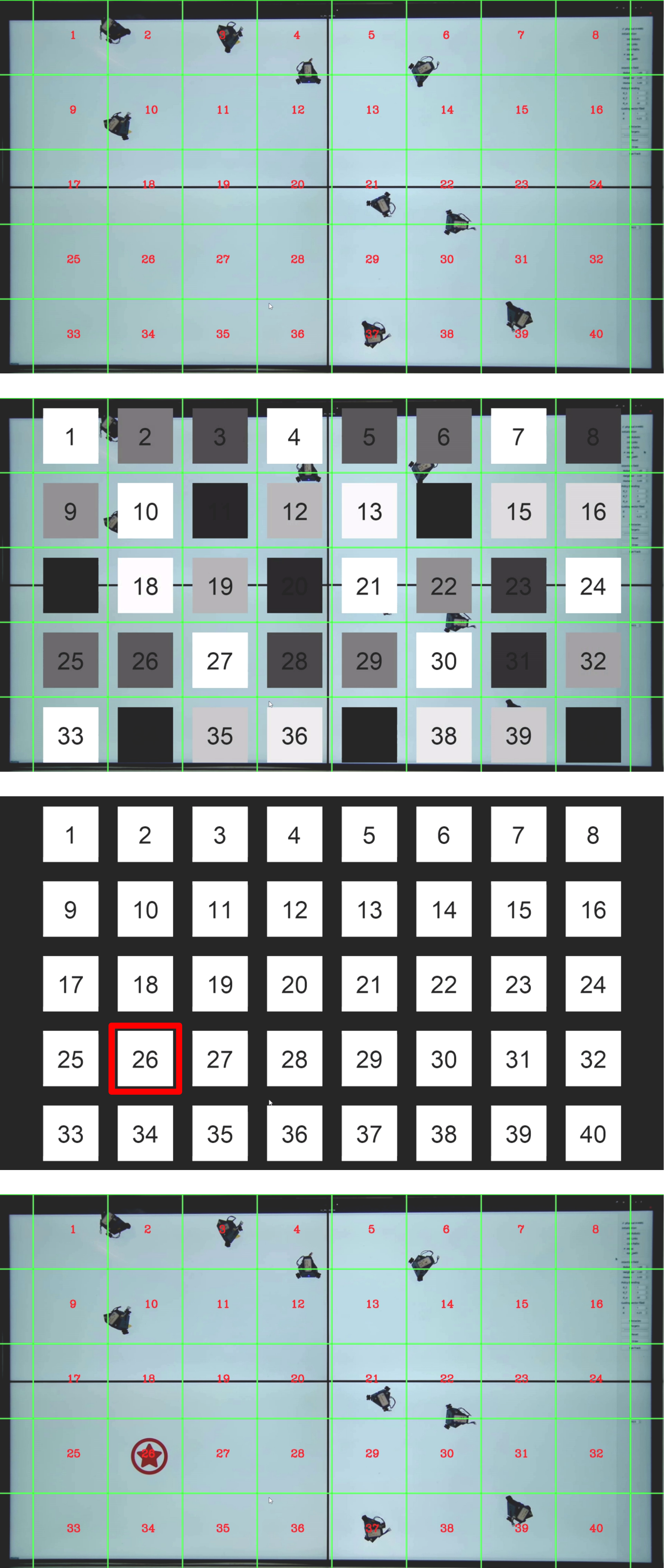}}
 \hspace{1pt}
	\subfloat[The process of robots moving toward the target]{\includegraphics[width=0.49\linewidth]{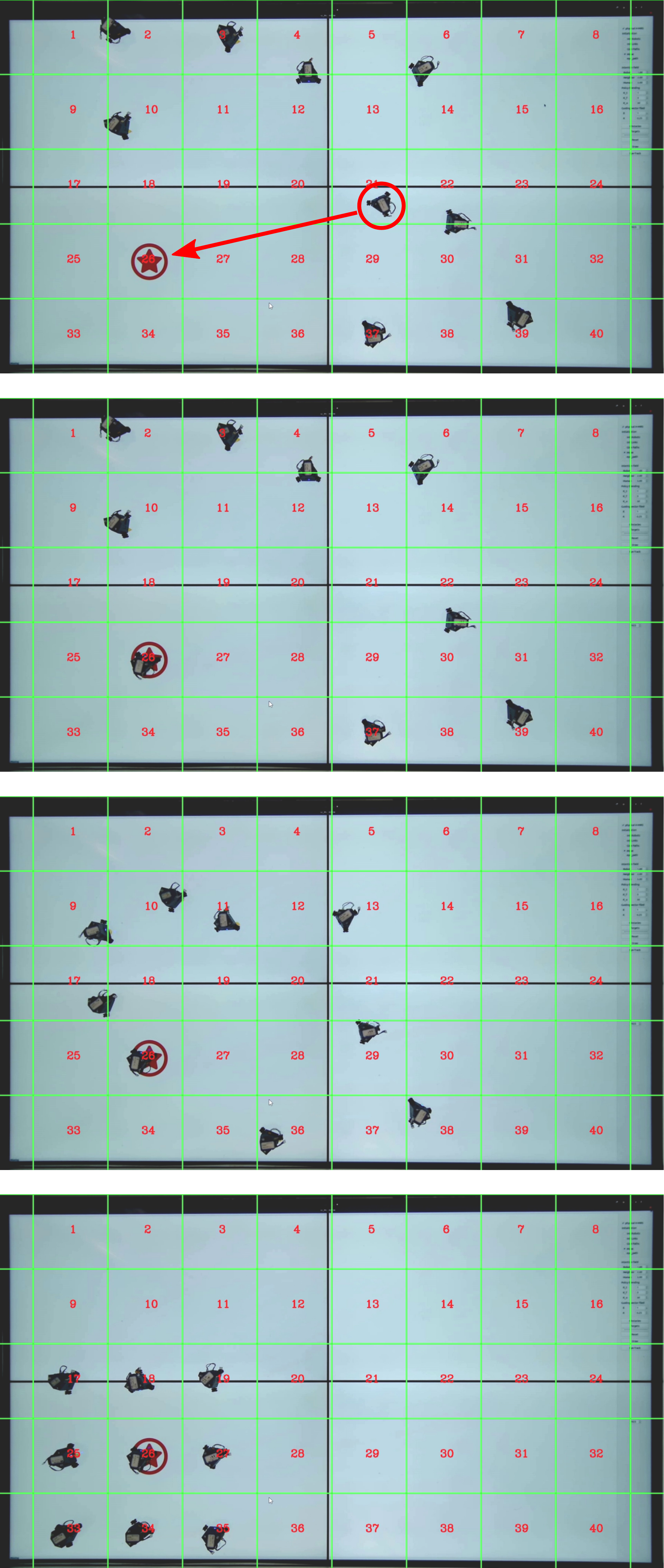}}
	\caption{Selecting targets by BCI.} 
	\label{bci_process}
\end{figure}

Although BCIs can obtain the corresponding results just through human imagination, this approach is not very accurate.
Therefore, we use a noninvasive BCI by visual stimulation to improve recognition accuracy.
Our experiments adopt the steady-state visual evoked potentials (SSVEP) paradigm, which typically has high information transfer rates, a simple system design, and few training sessions.
Further, the acquired EEG signals are decoded via the canonical correlation analysis (CCA) method.
As the most stable and versatile SSVEP EEG decoding algorithm, CCA statistically analyses the potential correlation of measured multidimensional variables. 
Without training or fine-tuning, it can directly map complex EEG signals to corresponding stimulus blocks.
The process of area selection is shown from top to bottom in Fig. \ref{bci_process}(a).
In our experiments, the image transmitted by the camera is uniformly segmented into 40 areas, as shown in the first figure in Fig. \ref{bci_process}(a). 
When the subject starts to select, each block will blink at a frequency of 8-16Hz with an interval of 0.2Hz to induce specific responses corresponding to the stimulus blocks, as shown in
the second figure in Fig. \ref{bci_process}(a). 
As an efficient spatial filtering algorithm, CCA calculates the correlation coefficients between EEG signals and targets of different frequencies and selects the stimulus block with the highest correlation coefficient as the subject's expected region.
% By gazing at the screen, the BCI can collect and process signals from the visual cortex of the brain, so the subject selects the corresponding area. 
In fact, the selection result for each area is in the form of a probability, except that the center region has the highest probability.
In Fig. \ref{bci_process}(b), area 26 is selected, and the target is placed in the corresponding position. 
For a simple verification of BCI control of multiple robots, the subject selects a certain target area online and guides multiple robots to reach around the target area, as shown from top to bottom in Fig. \ref{bci_process}(b).
The robots are first dispersed throughout the test field, and when the subject chooses area 26, the robot circled in red in the first figure moves to the target area first, followed by the remaining robots moving around the target.

% \begin{figure}[h!]
% 	\centering
% 	\subfloat[]{\includegraphics[width=0.8\linewidth]{figure/bci_sel1.pdf}}
%  \\
% 	\subfloat[]{\includegraphics[width=0.8\linewidth]{figure/bci_sel2.pdf}}
% 	\caption{The robot locomotion experiments.} 
% 	\label{bci}
% \end{figure}

\subsection{EMG for generating robot action commands}
\begin{figure}[h!]
	\centering
	\subfloat[upward]{\includegraphics[width=0.46\linewidth]{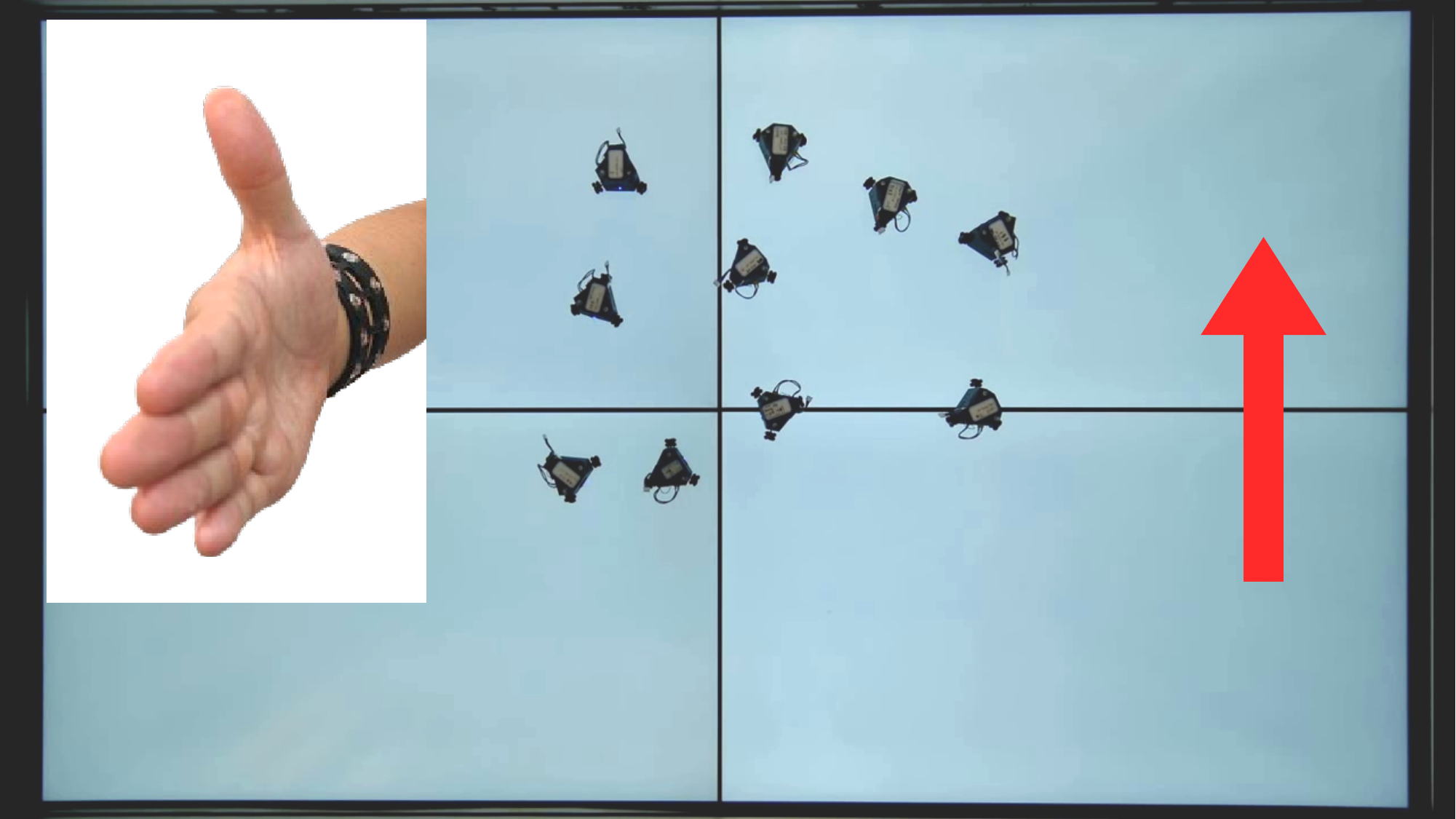}}
 \hspace{2mm}
	\subfloat[downward]{\includegraphics[width=0.46\linewidth]{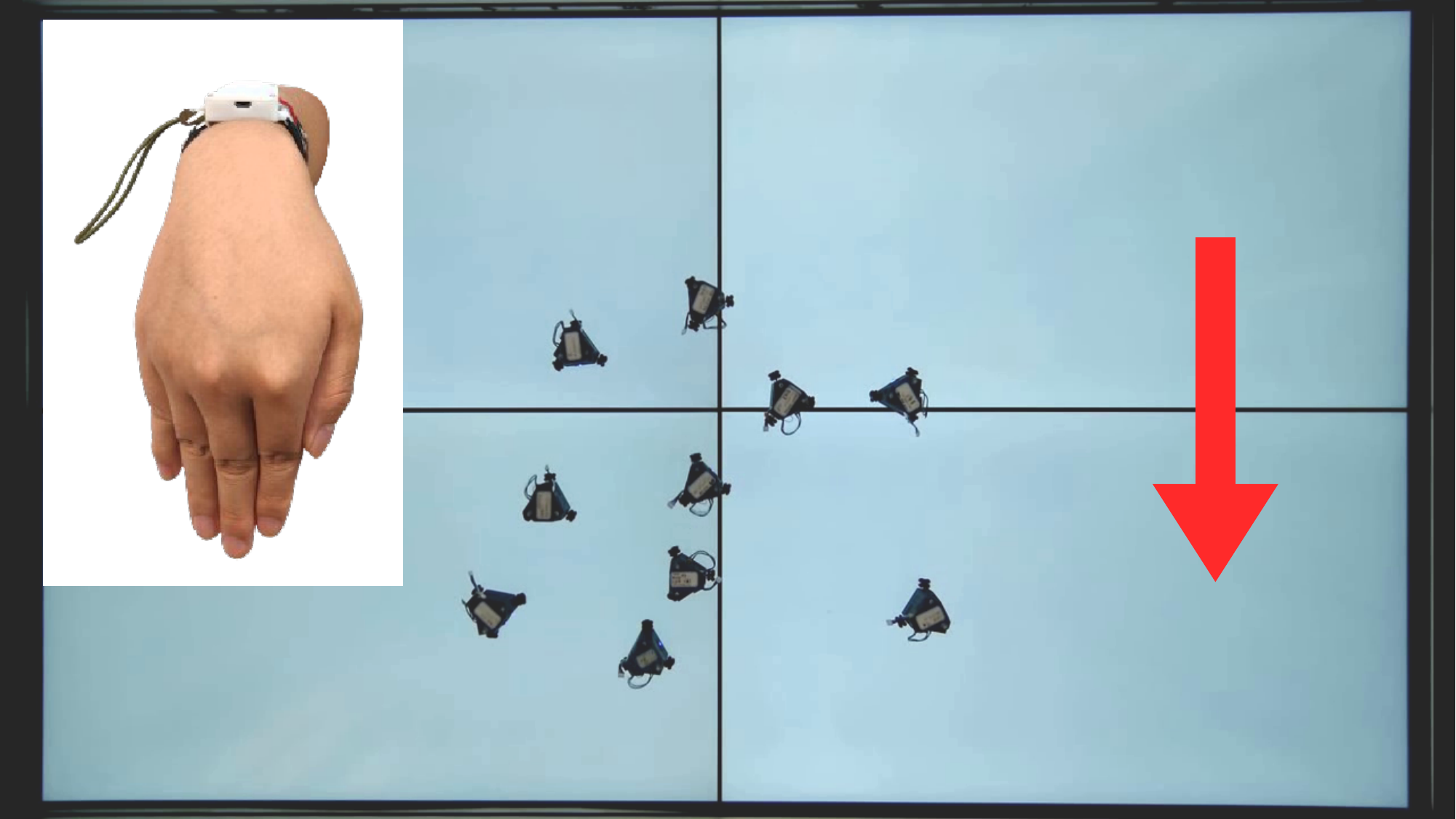}}
 \\
 	\subfloat[stop]{\includegraphics[width=0.46\linewidth]{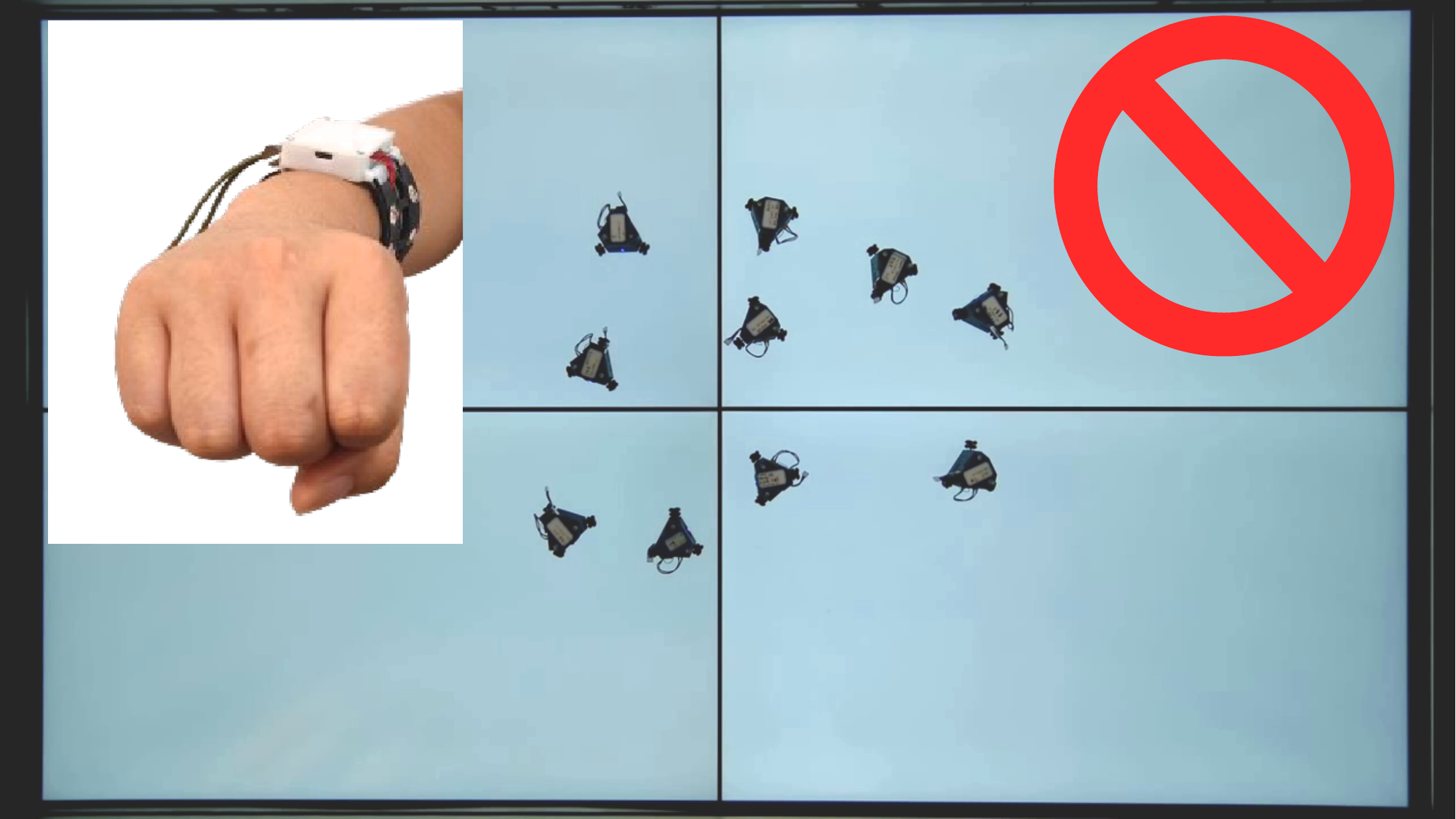}}
  \\
	\subfloat[left]{\includegraphics[width=0.46\linewidth]{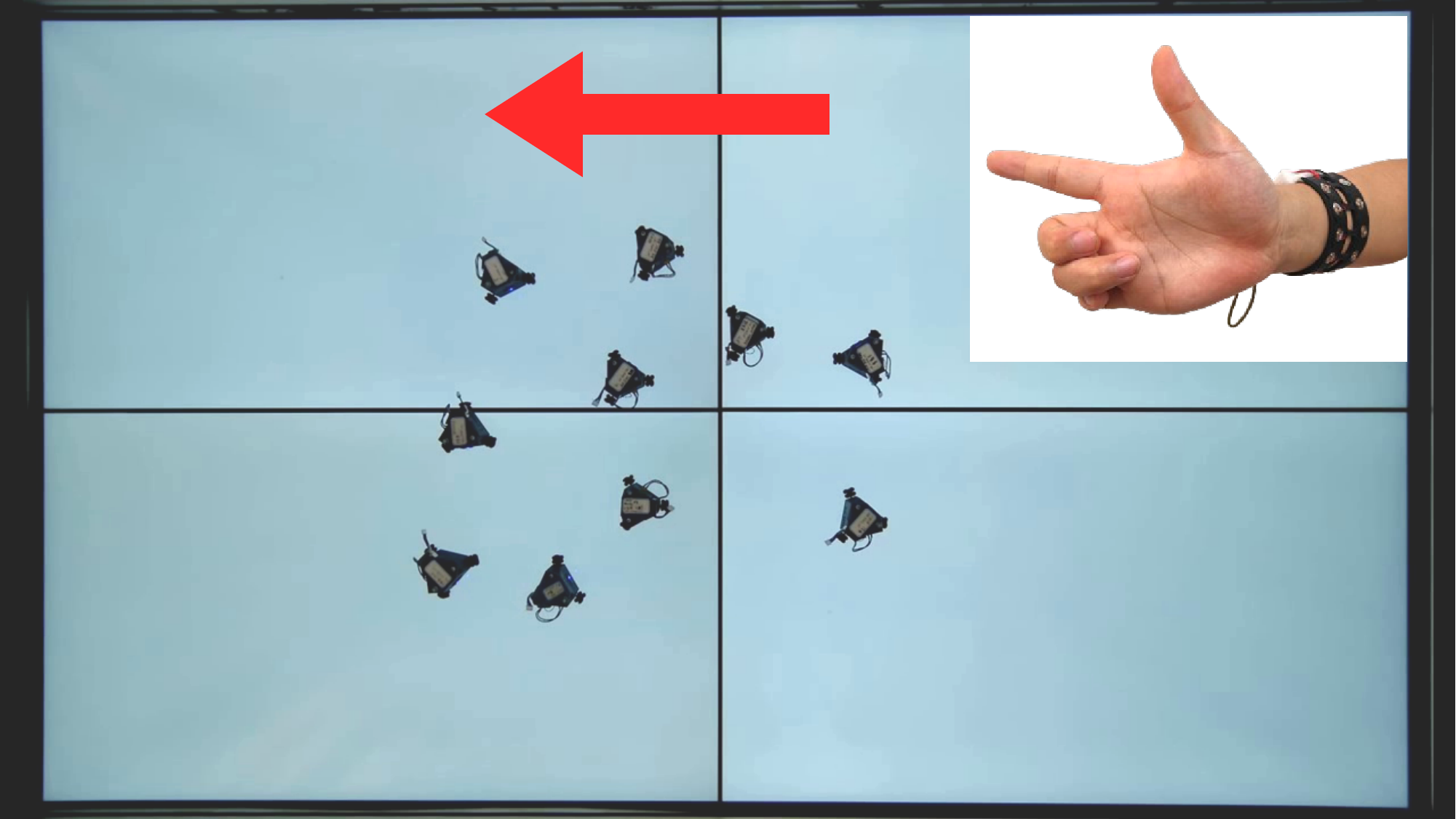}}
 \hspace{2mm}
 	\subfloat[right]{\includegraphics[width=0.46\linewidth]{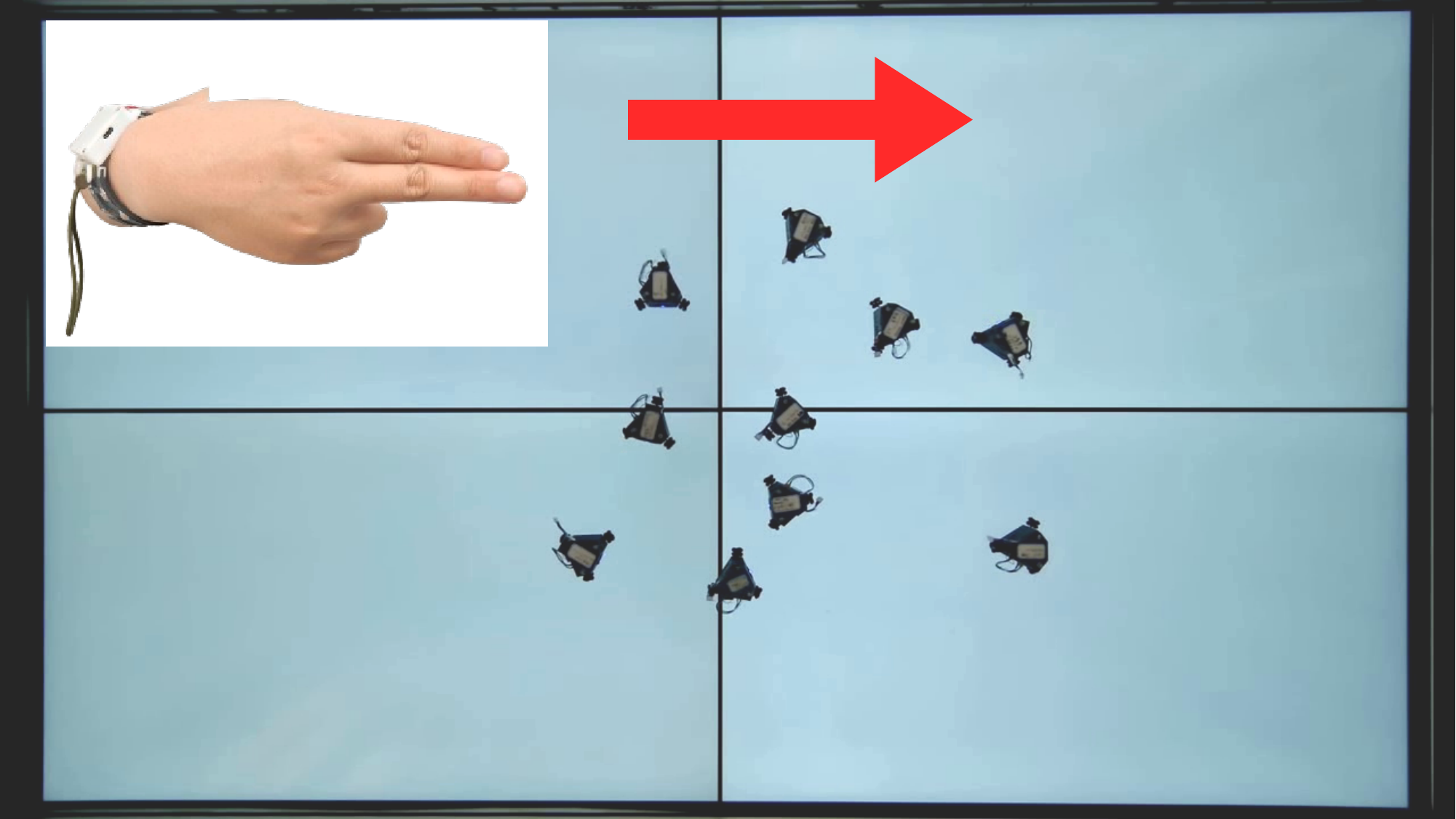}}
	\caption{The gestures control robot movement.} 
	\label{emg_ex}
\end{figure}

Through its capability to discern EMG signals, the EMG interface effectively senses gestural shifts, translating them into distinctive command outputs. We have devised five straightforward gestures, each of which can be reliably recognized, enhancing interaction intuitiveness.
As shown in Fig. \ref{emg_ex}, these five gestures instinctively correlate to five fundamental control commands: stop, upward motion, downward motion, left motion, and right motion, respectively.
These gestures have been meticulously designed to be as distinct from each other as possible, optimizing recognition accuracy and minimizing ambiguity during interpretation.
To facilitate this, we utilize the wristband, as illustrated in Fig. \ref{system framework}, for the acquisition of EMG data corresponding to these gestures, forming the basis of an offline dataset. In our experiments, a streamlined three-layer neural network is employed for gesture classification, ensuring efficiency without compromising precision.
Given the inherent variability of EMG signals across individuals, a brief calibration phase is necessary before experiments, during which subjects undergo a personalized network training session. To mitigate erroneous commands, 
a deliberate strategy is employed: gesture recognition must stabilize post-transition before transmitting new control signals to the robots. While this measure effectively reduces misidentification, it inevitably introduces a latency of about 0.5s in command issuance.
In our experiments, a single participant successfully controls ten robots to move simultaneously, as shown in Fig. \ref{emg_ex}.
% \begin{figure}[h!]
% 	\centering
% 	\subfloat[upward]{\includegraphics[width=0.28\linewidth]{figure/forward}}
%  \hspace{2mm}
% 	\subfloat[stop]{\includegraphics[width=0.28\linewidth]{figure/stop}}
%  \hspace{2mm}
%  	\subfloat[downward]{\includegraphics[width=0.28\linewidth]{figure/back}}
%   \\
% 	\subfloat[left]{\includegraphics[width=0.42\linewidth]{figure/left}}
%  \hspace{2mm}
%  	\subfloat[right]{\includegraphics[width=0.42\linewidth]{figure/right}}
% 	\caption{The designed gestures in the experiments.} 
% 	\label{gesture}
% \end{figure}

\subsection{Eye tracking for desired trajectory generation}
The eye-tracking technology captures the real-time gaze position on the screen, enabling intuitive and immediate transmission of intentions. It facilitates interactions where, for instance, directing one's gaze towards a specific point on the screen can instruct a robot to navigate to that location, or tracing out a path with one's gaze can effortlessly control the robot's travel route, which is challenging to achieve with BCI.

\begin{figure}[h!]
	\centering
	\includegraphics[width=0.935\linewidth]{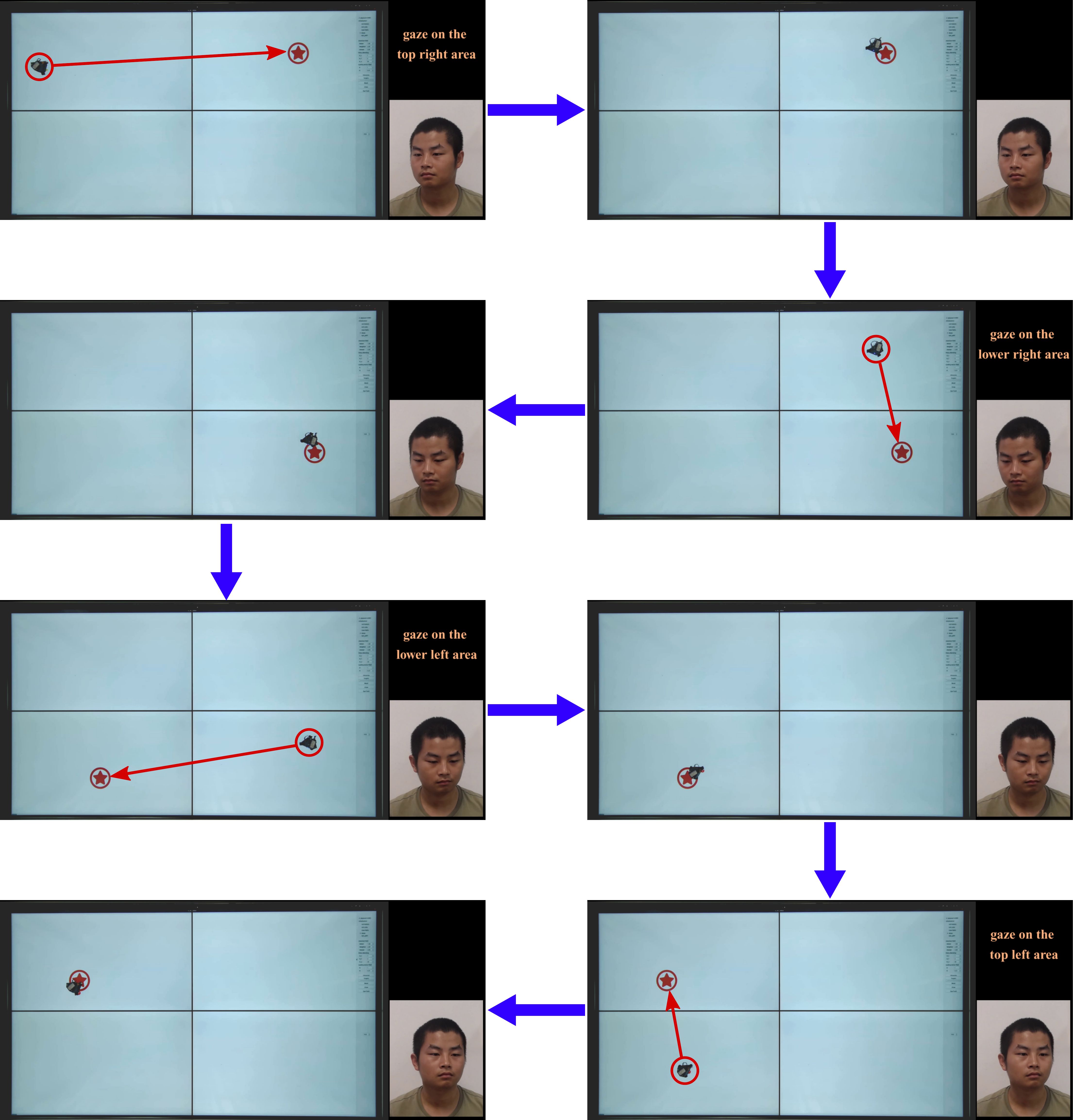}
      \caption{Guiding the robot to the target by gazing at a certain area.} 
	\label{eye_sig}
\end{figure}

\begin{figure}[h!]
	\centering
	\subfloat[Drawing a trajectory by eye movement]{\includegraphics[width=0.44\linewidth]{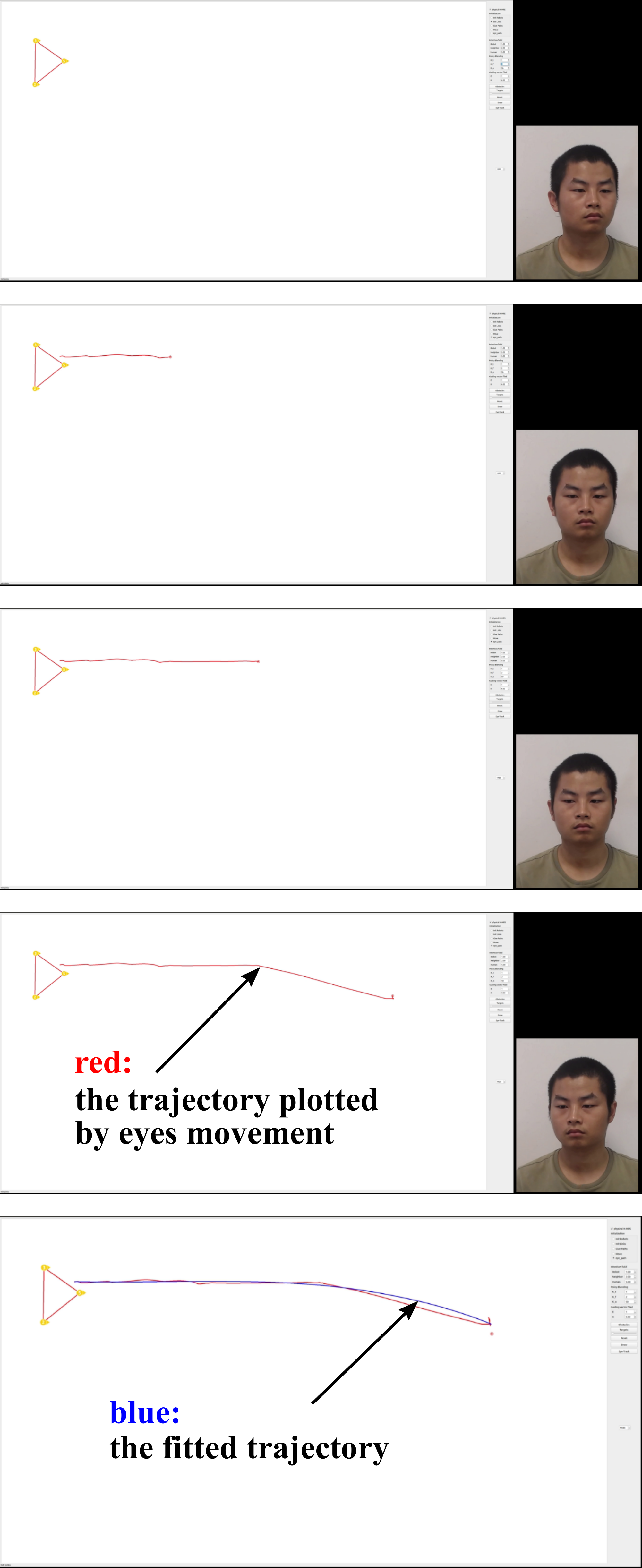}}
 \hspace{10pt}
	\subfloat[The formation's movement process]{\includegraphics[width=0.37\linewidth]{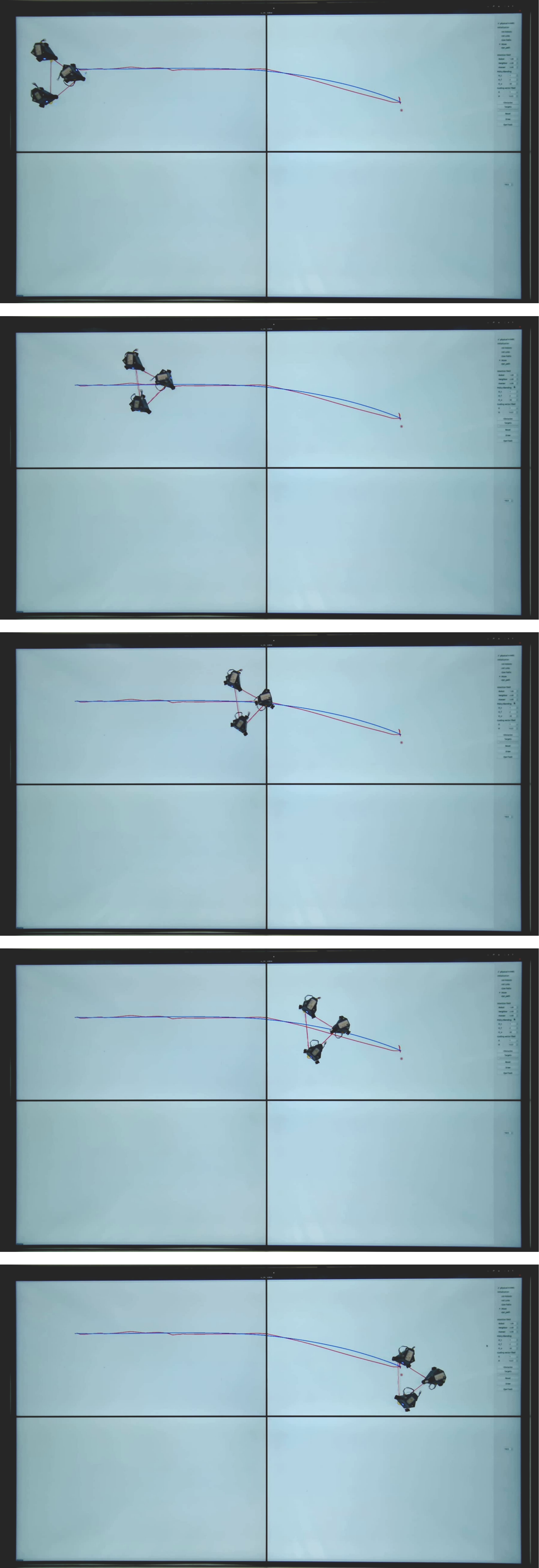}}
	\caption{Guiding the formation along a gaze-drawn trajectory.} 
	\label{eye_for_process}
\end{figure}

Before experiments, the eye-tracker should be affixed beneath the screen for positional calibration, and the distance between the subject and the screen should be 45-95 cm to guarantee optimal tracking performance.
Firstly, we use eye tracking to dynamically adjust the target position so that the robot keeps moving to the area where the subject is gazing, as shown in Fig. \ref{eye_sig}.
The red pentagram is the area the subject is gazing at, circled by the red circle is the robot, and the blue arrow represents the sequence of the experimental procedure. 
The participant sequentially directs their gaze to the upper right, lower right, lower left, and finally, upper left regions, prompting the robot to navigate accordingly in those respective directions.
Then, we use gaze movement to draw a continuous trajectory that guides the formation to move.
When the acquisition of gaze movement data begins, the subject can draw a trajectory with their gaze in the designed QT interface. A smooth trajectory is fitted when the acquisition ends, as shown from top to bottom in Fig. \ref{eye_for_process}(a).
When robots are given a command to move, the three robots can form a formation to move along the trajectory, as shown in Fig. \ref{eye_for_process}(b).

Through these human-robot interaction case studies, we have validated the capabilities of the DVRP-MHSI system. Leveraging an array of human input modalities, we test scenarios that demonstrate the successful human-robot collaboration. Acknowledging that the breadth of our experimental scenarios is circumscribed, it is pivotal to emphasize that the paramount contribution of our research resides in the development of the DVRP-MHSI itself.
These studies, while straightforward, demonstrate the system's capability to incorporate user inputs with the robots in real-time, fulfilling the design requirements presented in this paper.
A video of the experiment is available online\footnote{\href{https://www.bilibili.com/video/BV1LMepeiESM/}{https://www.bilibili.com/video/BV1LMepeiESM/}}. Moreover, the video of related multi-robot studies based on DVRP-MHSI (including multi-robot task allocation, multi-robot formation, multi-robot path planning, and multi-robot circumnavigation) is also available online\footnote{\href{https://www.bilibili.com/video/BV11Hp2e9EvY/}{https://www.bilibili.com/video/BV11Hp2e9EvY/}}.
% \subsubsection{multitouch for target area and trajectory} generation

\section{Conclusion and future work}
This work devises the Dynamic Visualization Research Platform for Multimodal Human-Swarm Interaction (DVRP-MHSI), an innovative platform dedicated to advancing research in human-swarm interaction. 
As the main testing ground, multi-touch screens can detect objects placed above them and dynamically render scenarios. 
This feature greatly facilitates the design and switching of mission scenarios.
As the platform's kinetic component, a small three-wheeled omnidirectional robot has been designed to be low-cost and easy to maintain, manufacture, and operate.
As a complement to the hardware, the software architecture of the system is designed to be highly flexible. It not only enables effortless customization of software components but also accommodates various human-computer interaction devices.
%This paper provides a brief survey of swarm simulators, swarm robotics platforms, and several human-swarm platforms.
%The survey of these existing swarm simulators and systems inspires the design idea of our proposed platform. 
%Namely, we
We use a simulated decentralized hardware approach to provide the required versatility for the DVRP-MHSI, enabling researchers to rapidly develop and deploy new algorithms. 
The developed platform is verified by utilizing various human-computer interaction devices to control a multi-robot system.

%In the future, we will develop variable and easy-to-deploy scenarios using Unity, thus enriching the human-robot interaction experience. We also aspire to exploit the potential of human engagement with robotic swarms, broadening the range of human-robot interaction interfaces. For instance, research can focus on multimodal intent fusion, which involves combining different intent modalities for a single user, or integrating intents from multiple users using the same or different interaction modalities. In addition, by analyzing biological signals such as electroencephalography and electromyography, our platform helps to explore how robot swarms can affect human cognitive processes and thus even initiate research related to rehabilitation robotics. Moreover, equipping the robots with an expanded sensor suite is envisaged to enhance their autonomy, further conducting intelligent, adaptive swarm robotics research.

\bibliographystyle{IEEEtran}
\bibliography{IEEEabrv,IEEEexample}

\end{document}